\documentclass[letterpaper]{article} % DO NOT CHANGE THIS
\usepackage{aaai25}  % DO NOT CHANGE THIS

\usepackage{ltl}
\usepackage{subcaption}
\usepackage{array}
\usepackage{float}
\usepackage{tikz}
\usepackage{xcolor}
\usepackage{xspace}
\usepackage{amsmath}
\usepackage{smartdiagram}
\usepackage{multirow}
\usepackage{booktabs}
\usepackage{pgfplots}
\usepackage{pgfplotstable}
\usepackage{filecontents}
\usepackage{amsmath,amssymb,amsfonts}
\usepackage{bm}

\usepackage{times}  % DO NOT CHANGE THIS
\usepackage{helvet}  % DO NOT CHANGE THIS
\usepackage{courier}  % DO NOT CHANGE THIS
\usepackage[hyphens]{url}  % DO NOT CHANGE THIS
\usepackage{graphicx} % DO NOT CHANGE THIS
\usepackage{natbib}  % DO NOT CHANGE THIS AND DO NOT ADD ANY OPTIONS TO IT
\usepackage{caption} % DO NOT CHANGE THIS AND DO NOT ADD ANY OPTIONS TO IT
\usepackage{algorithm}
\usepackage{algorithmic}

\usepackage{newfloat}
\usepackage{listings}
\DeclareCaptionStyle{ruled}{labelfont=normalfont,labelsep=colon,strut=off} % DO NOT CHANGE THIS
\floatstyle{ruled}
\newfloat{listing}{tb}{lst}{}
\floatname{listing}{Listing}
\usetikzlibrary{matrix}
\usetikzlibrary{automata}
\usetikzlibrary{positioning}
\usetikzlibrary{shapes.geometric, arrows}
\usetikzlibrary{shadows.blur}
\usetikzlibrary{tikzmark}
\usetikzlibrary{calc}
\tikzset{
  -|-/.style={->,draw,
    to path={
      (\tikztostart) -| ($(\tikztostart)!#1!(\tikztotarget)$) |- (\tikztotarget)
      \tikztonodes
    }
  },
  -|-/.default=0.5,
  |-|/.style={
    to path={
      (\tikztostart) |- ($(\tikztostart)!#1!(\tikztotarget)$) -| (\tikztotarget)
      \tikztonodes
    }
  },
  |-|/.default=0.5,
}
\tikzset{
        node distance=4cm, % specifies the minimum distance between two nodes. Change if n
        every state/.style={minimum size=5pt, thick, fill=gray!10}, % sets the properties for each ’state’ n
        initial text=$ $, % sets the text that appears on the start arrow
    }

\newcommand{\upd}[2]{\ensuremath{[ \hspace{0.5pt} #1 \leftarrowtail \, #2 \hspace{0.5pt} ]}}
\newcommand{\name}[1]{{\text{\texttt{#1}}}}

\newcommand{\term}{\ensuremath{\tau}}
\newcommand{\fterm}{\ensuremath{\term_{F}}\xspace}
\newcommand{\pterm}{\ensuremath{\term_{P}}\xspace}

\newcommand{\terms}{\ensuremath{\mathcal{T}}\xspace}
\newcommand{\pterms}{\ensuremath{\terms_{\!P\hspace{-0.5pt}}}\xspace}
\newcommand{\fterms}{\ensuremath{\terms_{\!F\hspace{-0.5pt}}}\xspace}

\newcommand{\sep}{\ensuremath{\quad | \quad}}

\newcommand{\sats}{\ensuremath{\;\vDash_{\!\langle \hspace{-1pt} \cdot \hspace{-1pt} \rangle}}}

\newcommand{\branch}[2]{\ensuremath{#1 \! \wr \hspace{-0.6pt} #2}}
\newcommand{\assign}[1]{\ensuremath{\langle #1 \rangle}}
\newcommand{\functions}{\ensuremath{\mathcal{F}}\xspace}
\newcommand{\fnames}{\ensuremath{\mathbb{F}}\xspace}

\lstdefinelanguage{JavaScript}{
  morekeywords=[1]{break, continue, delete, else, for, function, if, in,
    new, return, this, typeof, var, void, while, with},
  morekeywords=[2]{false, null, true, boolean, number, undefined,
    Array, Boolean, Date, Math, Number, String, Object},
  morekeywords=[3]{eval, parseInt, parseFloat, escape, unescape},
  sensitive,
  morecomment=[s]{/*}{*/},
  morecomment=[l]//,
  morecomment=[s]{/**}{*/}, % JavaDoc style comments
  morestring=[b]',
  morestring=[b]"
}[keywords, comments, strings]

\definecolor{mediumgray}{rgb}{0.3, 0.4, 0.4}
\definecolor{mediumblue}{rgb}{0.0, 0.0, 0.8}
\definecolor{forestgreen}{rgb}{0.13, 0.55, 0.13}
\definecolor{darkviolet}{rgb}{0.58, 0.0, 0.83}
\definecolor{royalblue}{rgb}{0.25, 0.41, 0.88}
\definecolor{crimson}{rgb}{0.86, 0.8, 0.24}

\lstdefinestyle{JSES6Base}{
  backgroundcolor=\color{white},
  basicstyle=\ttfamily \small,
  breakatwhitespace=false,
  breaklines=false,
  captionpos=b,
  columns=fullflexible,
  commentstyle=\color{mediumgray}\upshape,
  emph={},
  emphstyle=\color{crimson},
  extendedchars=true,  % requires inputenc
  fontadjust=true,
  frame=single,
  identifierstyle=\color{black},
  keepspaces=true,
  keywordstyle=\color{mediumblue},
  keywordstyle={[2]\color{darkviolet}},
  keywordstyle={[3]\color{royalblue}},
  numbers=left,
  numbersep=5pt,
  numberstyle=\tiny\color{black},
  rulecolor=\color{black},
  showlines=true,
  showspaces=false,
  showstringspaces=false,
  showtabs=false,
  stringstyle=\color{forestgreen},
  tabsize=2,
  title=\lstname,
  upquote=true  % requires textcomp
}

\lstdefinestyle{JavaScript}{
  language=JavaScript,
  style=JSES6Base
}
\lstdefinestyle{ES6}{
  language=ES6,
  style=JSES6Base
}

\begin{document}
% \linenumbers
\setcounter{secnumdepth}{2}

\title{Procedural Adherence and Interpretability Through Neuro-Symbolic Generative Agents}

% \author{Anonymous}
\author {
    % Authors
    Raven Rothkopf,
    Hannah Tongxin Zeng,
    Mark Santolucito
}
\affiliations {
    % Affiliations
    Barnard College, Columbia University\\
    \{rgr2124, tz2486, msantolu\}@barnard.edu
}
% \author{
%     \IEEEauthorblockN{Raven Rothkopf}
%     \IEEEauthorblockA{\textit{rgr2124@barnard.edu} \\
%     \textit{0000-0002-3926-683X}}
%     \and
%     \IEEEauthorblockN{Hannah Tongxin Zeng}
%     \IEEEauthorblockA{\textit{tz2486@barnard.edu} \\
%     \textit{} \\
%     \\ Barnard College, Columbia University \\ New York, USA
%      }
%     \and
%     \IEEEauthorblockN{Mark Santolucito}
%     \IEEEauthorblockA{\textit{msantolu@barnard.edu} \\
%     \textit{0000-0001-8646-4364}}
% }

\maketitle 

\begin{abstract}
The surge in popularity of large language models (LLMs) has opened doors for new approaches to the creation of interactive agents. 
However, managing and interpreting the temporal behavior of such agents over the course of a potentially infinite interaction remain challenging. 
The stateful, long-term horizon reasoning required for coherent agent behavior does not fit well into the LLM paradigm. 
We propose a combination of formal logic-based program synthesis and LLM content generation to bring guarantees of procedural adherence and interpretability to generative agent behavior. 
To illustrate the benefit of procedural adherence and interpretability, we use Temporal Stream Logic (TSL) to generate an automaton that enforces an interpretable, high-level temporal structure on an agent.
With the automaton tracking the context of the interaction and making decisions to guide the conversation accordingly, we can drive content generation in a way that allows the LLM to focus on a shorter context window.
% With TSL, we can augment generative agents so that developers have a higher level of guarantees on behavior and better interpretability of their system.
We evaluated our approach on different tasks involved in creating an interactive agent specialized for generating choose-your-own-adventure games. We found that over all of the tasks, an automaton-enhanced agent with procedural guarantees achieves at least 96\% adherence to its temporal constraints, whereas a purely LLM-based agent demonstrates as low as 14.67\% adherence.
\end{abstract}

% three different domains: a choose-your-own-adventure game generator, a math textbook aid that students can consult while reading, and conversation manager that constrains the dialogue between two autonomous agents.

% \keywords{Program Synthesis, Large Language Models, Explainable AI, Temporal Stream Logic, Human-AI Co-creation}

\begin{figure*}[h]
\centering
\includegraphics[scale=0.95]{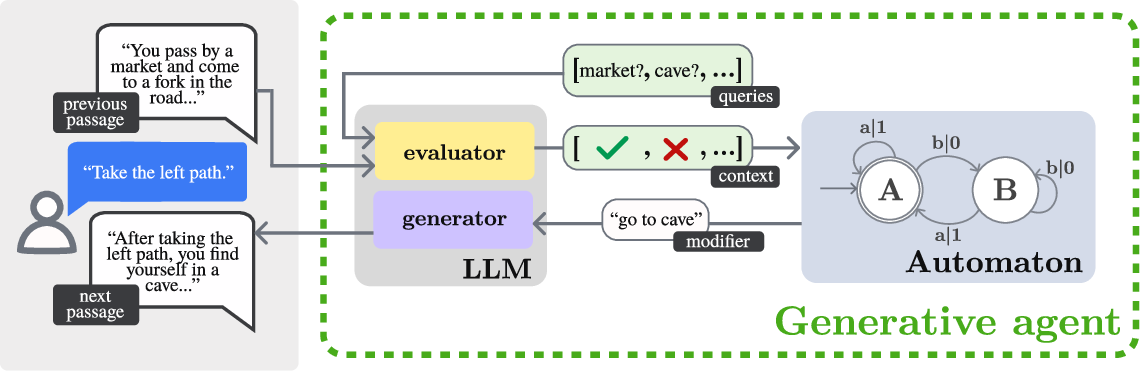}
\caption{An example turn between an end-user and a choose-your-own-adventure game agent. After receiving a user prompt, the LLM consults an internal automaton synthesized from the Temporal Stream Logic (TSL) specification in Fig.~\ref{fig:adventure}. Using previous story context, the automaton outputs a correct prompt modifier so that the agent's response aligns with the story's formal specification of long-horizon behavior. See Secs.~\ref{sec:agent} and~\ref{sec:system} for an in-depth explanation of the agent, Sec.~\ref{sec:correctness} for formal descriptions of the agent's procedural adherence and interpretability, and Sec.~\ref{sec:eval} for a preliminary agent evaluation.}
\label{fig:overview}
\end{figure*}

\section{Introduction}
%what is the high level context and motivation
Large language models (LLMs) excel as generative agents that should have some self-consistency in their behavior.
One such example is the use of LLMs to create video game agents~\cite{latouche2023generating,croissant2023appraisal}. 
However, one limitation is in the ability to manage long-horizon behavioral consistency of such agents. Often, when relying solely on LLMs to control agent behavior, the agent will display behavior that is out of line with its previously established patterns~\cite{binz2023using}.
While a certain amount of agent evolution can increase the depth of player experience~\cite{beale2009affective}, players also expect in-game agents to have some consistency over time~\cite{gilleade2005affective}.
Prior work has attempted to address the issues of agent memory by logging an input-response history which can be used in future calls to a LLM for content (e.g. NPC script). 
While these approaches create more natural conversation flows, e.g. with chain-of-thought frameworks~\cite{wei2022chain}, the core problem of enforcing temporal consistency remains unaddressed.

%what has existing work identified as an issue in this space
Enforcing temporal consistency for our work means to give structure to the long-horizon behavior of a pre-trained LLM.
For example, we may want to ensure that an LLM teaching assistant for a CS101 class does not give code with nested conditionals before explaining what a conditional is and how it functions.
To impose such constraints on an LLM agent, developers must compose extensive prompts, scrutinize responses for potential factual inaccuracies, biases, or content misaligned with their intentions, adjust their prompts accordingly, and repeat the process. 
This creates the ``Gulf of Execution"~\cite{hutchins1985direct}, representing the divide between developers' intentions and the system's ability to fulfill them, leading to excessive back-and-forth in crafting prompts for developers to guide LLMs toward desired outputs.
In addition, the ``Gulf of Evaluation" exists between the system's output and developers' comprehension of how well these outputs align with their intentions.
This lack of interpretability of the model necessitates additional effort from the developer to understand how and why the LLM-based agent came up with its responses. 

%how will we solve this
To bridge these two gulfs and to enforce temporally consistent agent behavior in LLMs, we leverage temporal logic and program synthesis.
We combine LLMs with logic-based program synthesis techniques to build a framework for creating generative agents that are guaranteed to obey particular temporal properties of the generative content.
In this system, developers provide a specification of the temporal behavior to which their agent should adhere, and we synthesize a system that structures LLM queries of the agent.

%what are the design goals of our solution
The key design goal of our approach is an interactive generative agent with two main features: 
\begin{enumerate}
    \item  \textbf{Adherence}: Despite explicit instructions in prompts, the LLM might still generate content that clearly violates the instructions~\cite{mu2023can}. We want to make sure that the agent adheres to the provided specification.
    \item  \textbf{Interpretability}: When an LLM produces unexpected outcomes, it is difficult to pinpoint the cause from previous interactions and correct accordingly~\cite{Liao2024AI}. The developer should be able to identify the root cause of decisions made by the agent.
    % \item \textbf{Modularity}: LLMs fall short of representing a non-linear information structure that involves more complex components such as conditionals and parallelism. This is particularly an issue for enabling the iterative creation developers require for building complex systems.
\end{enumerate}
% --this challenge has been termed \textit{loss of authorial control}~\cite{mu2023can}
The developers of the generative agent are the primary users of our system, as opposed to the end-users interacting with the generative agent in post-production. 
%\markk{can we cite something here about prompt engineering and trying to get an LLM to do two things once is hard}

% how we do solve the key problem
Recognizing these design priorities and that generative agents are reactive systems that consume input and produce output while maintaining an internal state, we turn to reactive synthesis.
Reactive synthesis can take a specification of an agent's long-horizon behavior and synthesize an automaton that gives instructions (as a prompt modifier) to the LLM of how to respond on each state transition.
To make decisions on how to transition between states, the automaton gets LLM queries on the previous LLM output at each turn.

Temporal Stream Logic (TSL)~\cite{finkbeiner2019temporal} is one approach to specifying an agent's long-horizon behavior. 
With TSL, we write properties about the temporal behavior of an agent, then synthesize an automaton that is guaranteed to meet those properties in program code. 
For example, an agent specialized to generate text for interactive choose-your-own adventure games is shown in Fig.~\ref{fig:overview}. 
This particular agent uses an internal automaton from the TSL specification in Fig.~\ref{fig:adventure} to ensure the player visits a cave only after visiting a market.
There are no guarantees about the behavior of the component functions in Fig.~\ref{fig:adventure}, but using TSL, we can construct an automaton that governs the times when these functions are called. 
The arc of the story and how the player navigates the story change over time, but their temporal relationship remains consistent.

With this hybrid automaton-LLM system, we can treat a turn--or a prompt and agent's response--as two different tasks: generating a correct response according to the prompt, which is controlled by the LLM, and choosing the correct prompt modifier that aligns with the agent's long-horizon behavior using the context of previous turns, which is controlled by the automaton. This approach distinguishes between \textbf{point-in-time} (low-level agent behavior over one turn) vs. \textbf{procedural} (high-level agent behavior over an entire conversation) adherence and interpretability.

Leveraging the correct-by-construction automaton, we get a generative agent with guaranteed procedural adherence and preexisting theories to achieve procedural interpretability for free. We obtain significantly better long-horizon results with our automaton-enhanced agent than state-of-the-art approaches that only rely on an LLM.

Our goal is to illustrate and evaluate one of \emph{many} possible ways to guarantee procedural adherence and interpretability of an LLM. We highlight the importance of the distinction between and separation of procedural and point-in-time properties when building neuro-symbolic generative agents. 

The key contributions of this work are as follows:
\begin{enumerate}
    \item A formalization of generative agents as reactive systems and notions of procedural adherence and interpretability.
    \item A prototype system using Temporal Stream Logic (TSL) to demonstrate one approach to gaining procedural adherence and interpretability in generative agent behavior.
    \item An evaluation of this framework measuring temporal consistency compared to existing chain-of-thought memory models in a choose-your-own-adventure agent, and an explanation of its interpretability.
\end{enumerate}

\begin{figure}[t!]
\begin{flushleft}
\small
\texttt{\textbf{ASSUME}}

\textit{//Once the LLM is queried for a story passage where the player visits a cave, in the next step, the player should be in a cave.}
\vspace{-0.2em}
\begin{flalign*}
\LTLglobally & \left([\texttt{storyPassage} \leftarrow \texttt{toCave(summary)}] \LTLequivalent \right. & \\
& \left. \LTLnext \texttt{inCave(summary)}\right) &
\end{flalign*}

\textit{//Once the LLM is queried for a story passage where the player visits a market, eventually the player should be in a market.}
\vspace{-0.2em}
\begin{flalign*}
\LTLglobally & \left([\texttt{storyPassage} \leftarrow \texttt{toMarket(summary)}]  \LTLimplication \right. & \\
& \left. \LTLfinally \texttt{inMarket(summary)}\right) &
\end{flalign*}
% \textit{//Once the LLM is queried for a story passage where the player visits a town, eventually the player should be in a town.}
% \vspace{-0.2em}
% \begin{flalign*}
% &\LTLglobally \left([\texttt{storyPassage} \leftarrow \texttt{toTown(summary)}] \LTLimplication \LTLfinally \texttt{inTown(summary)}\right) &
% \end{flalign*}
\texttt{\textbf{GUARANTEE}}

\textit{//The player cannot go to cave until they have been in a market.}
\vspace{-0.2em}
\begin{flalign*}
&\neg \texttt{inCave(summary)} \LTLweakuntil \texttt{inMarket(summary)} 
% \ \land &\\
% & \neg \texttt{inCave(summary)} \LTLweakuntil \texttt{inTown(summary)} 
&
\end{flalign*}
\textit{//Infinitely often, the player should visit the cave and the market.}
\vspace{-0.2em}
\begin{flalign*}
& \LTLglobally(\LTLfinally \texttt{toCave(summary)} \land \LTLfinally \texttt{toMarket(summary)}) 
% \land \LTLfinally \texttt{toTown(summary)})
&
\end{flalign*}
\end{flushleft}
\caption{A TSL spec (cf. Sec.~\ref{sec:synthesis}) constraining the locations the player visits in a choose-your-own-adventure game.} 
\label{fig:adventure}
\end{figure}

\section{Related Work}
\subsection{LLM Adherence and Interpretability}
Hallucinations in large language model (LLM) outputs--namely, plausible yet ungrounded information~\cite{dhuliawala2023chainofverification}--can lead to inconsistency and factual errors in generated conversations \cite{camburu2019make,austin2021program}. 
Verification methods, whether executed and planned internally by the LLM (referred to as self-critique)~\cite{press2022measuring,manakul2023selfcheckgpt,dhuliawala2023chainofverification}, coordinated among different LLMs to find agreement~\cite{du2023improving}, or completed outside the LLM framework entirely~\cite{min2023factscore}, contribute diverse approaches to addressing hallucinations. 
Enhancing the reasoning capacity of LLMs is another way to reduce hallucinations. 
Our work can be understood as adding a reasoning component to LLMs for verifying temporal consistency in its outputs. 
Compared to previous work that enhances reasoning ability in specific domains, such as reading comprehension~\cite{andor2019giving} and mathematics~\cite{geva2020injecting,pikekos2021measuring}, our work aims to facilitate reasoning of LLMs in the temporal domain, irrespective of the nature of tasks.

Explainabile AI (XAI), or understanding how an AI model arrives at its results, is an increasingly important field in the age of LLMs~\cite{zhao2024explainability}. Some approaches to explanation measure the relevance of each input feature to a model's prediction using methods like SHAP and integrated gradients~\cite{lundberg2017unified, sundararajan2017axiomatic}, analyze attention weights to interpret the model's focus~\cite{vig2019multiscale, hoover2019exbert}, generate textual descriptions to clarify decisions~\cite{camburu2018snli}, and reverse-engineer a model's inner workings to uncover causal mechanisms from inputs to outputs~\cite{elhage2021mathematical}. 
These methods focus on explaining a single turn between the LLM and an end-user, and fall short in explaining the cause of an LLM's response when looking back over a potentially infinite history of interactions influencing that response. 
Many previous XAI techniques are challenging to apply to LLMs due to their massive scale and proprietary internals~\cite{wei2023larger}. 

\subsection{Formal methods and LLMs}
Leveraging formal methods like specification languages to constrain generative agents is an area of critical importance~\cite{cohen2024survey}. 
Current approaches aim to enforce formally correct and transparent architectures on top of pre-trained LLMs, similar to our method. 
For instance, the Ceptre language uses liner logic to rapidly prototype game mechanics and generative narrative structures~\cite{martens2015ceptre}, and Linear Temporal Logic (LTL) automata have been used to encode safety constraints for robotic LLM agents~\cite{yang2023plug}. 
More recently, directed acyclic graphs have been employed to organize LLMs for scalable multi-agent collaboration~\cite{qian2024scaling}. 
We focus on generalizing and formalizing the procedural adherence and interpretability found in related techniques and our own implementation. 
\section{Preliminaries}

\subsubsection{Reactive Synthesis}\label{sec:synthesis}

While the framework of procedural adherence and interpretability can be applied to other formal models, the use of reactive synthesis~\cite{Church} is the core of our implementation and formalization.  
Reactive synthesis focuses on the temporally infinite interaction between the system and its environment, and automatically constructs control strategies that ensure that the system reacts to any possible move by the environment in a way that adheres to the specification's requirements. 
In classical synthesis, using specification languages such as Linear Temporal Logic~\cite{pnueli1977temporal}, time is discrete and inputs and outputs are given as vectors of Boolean signals. 
\emph{Temporal Stream Logic (TSL)}~\cite{finkbeiner2019temporal} is a specification language that provides an extra layer of abstraction that can be applied to input and output values of arbitrary type (as opposed to only Boolean as in LTL). 
Specifically, TSL introduces predicate terms, $ \pterm \in \pterms $, which are used to make observations on the environment, and function terms, $ \fterm \in \fterms $, which are used to construct output values.
Additionally, TSL introduces the notion of cell values--values output by the system and piped back as input in the following timestep.
The grammar of a TSL formula, $\varphi$, is:
\begin{equation*}
  \begin{array}{rl}
  \pterm \ \; ::= & \name{p}~\;\fterm^{0}~\;\fterm^{1}~\;\ldots~\;\fterm^{n-1} \\[0.3em]
  \fterm \ \; ::= & \name{s}_{\name{i}} \sep \name{f}~\;\fterm^{0}~\;\fterm^{1}~\;\cdots~\;\fterm^{n-1} \\[0.3em]
  \varphi \ \; ::= & \; \pterm \ \, | \ \, \upd{\name{s}_{\name{o}}}{\fterm} \ \, | \ \, \neg \varphi \ \, | \ \, \varphi \wedge \varphi \ \, | \ \, \LTLnext \varphi \ \, | \ \, \varphi \LTLuntil \varphi
  \end{array}
\end{equation*}
TSL is based on the usual LTL operators \emph{next} $\LTLnext$ and \emph{until} $\LTLuntil$. TSL also uses the standard derived operators, such as
\emph{release} $\varphi \LTLrelease \psi \equiv \neg(\neg\varphi \LTLuntil \neg\psi)$,
\emph{always} $\LTLglobally \varphi \equiv \bot \LTLrelease \varphi$,
\emph{eventually}  $\LTLeventually \varphi \equiv true \LTLuntil \varphi$, and 
\emph{weak until} $\varphi \LTLweakuntil \psi \equiv (\varphi \LTLuntil \psi) \lor (\LTLglobally \varphi)$.

TSL operates over input signals, $ \name{s}_{\name{i}} \in \mathbb{I} \cup \mathbb{C}$ which can be a stream of inputs or cell (looped) values, and output signals $ \name{s}_{\name{o}} \in \mathbb{O} \cup \mathbb{C} $, which can be a stream of outputs or cell values. All the available predicate names $\name{p}$ and all the available function names $\name{f}$ form the set of function symbols $\fnames$.
A TSL formula describes a system that consumes input $\mathcal{I} = \mathbb{I} \cup \mathbb{C}$ and produces output $\mathcal{O} = \mathbb{O} \cup \mathbb{C}$.

The realizability problem of TSL is stated as: given a TSL formula~$ \varphi $, is there
a strategy~$ \sigma \in \mathcal{I}^{+} \to \mathcal{O} $ mapping a finite input stream (since the beginning of time) to an output (at each particular time-step),
such that for any infinite input stream $ \iota \in \mathcal{I}^{\omega} $,
and every possible interpretation of the function symbols (some concrete implementation) $ \assign{\cdot} : \fnames \to \functions $, 
the execution of that strategy over the input $ \branch{\sigma}{\iota} $ satisfies $ \varphi $:
\begin{equation*}
  \exists \sigma \in \mathcal{I}^{+} \to \mathcal{O}. \ \, 
  \forall \iota \in \mathcal{I}^{\hspace{0.2pt}\omega}. \ \, 
  \forall \assign{\cdot} : \fnames \to \functions. \ \, 
  \branch{\sigma}{\iota}, \iota \sats \varphi
\end{equation*}
If such a strategy~$ \sigma $ exists, we say that $ \sigma $ realizes $ \varphi $.
The key insight here is that in TSL we specify a temporal relation of predicate evaluations to function applications--abstracting away from what these predicates and functions actually do to any underlying data.
In TSL synthesis, this model $\sigma$ can be turned into a block of program code that describes a Mealy machine~\cite{mealy1955method}, where the transitions represent function and predicate terms.
%For a full exposition of the formalism of TSL we refer the reader to~\cite{finkbeiner2019temporal}. 

\section{Generative Agents}\label{sec:agent}

The goal of this work is to build correct-by-construction generative agents. 
To ground our notion of correctness, we need a formal definition of a generative agent.
We are interested in generative agents that take actions at discrete time-steps as a reaction to observations on their environment and the history of prior environmental observations.
In this work we focus on a model where the environment and actions are text-based, though our approach is not specific to this (any datatype, e.g. JSON objects, would be fine).

This model of a generative agent matches closely with that of a reactive system, specifically with a layer of data abstraction as modeled by TSL.
In our setting, a generative agent, $\mathcal{A}$, is a reactive system which first consumes a stream of inputs $\mathcal{I}^{+}$ in the form of environmental text inputs (inputs $\mathbb{I}$) and its own prior outputs (cells $\mathbb{C}$).
At each timestep, the generative agent may make queries about these inputs (predicate terms \name{p}).
In our system, these  predicates are LLM queries (e.g. \texttt{inCave(summary)}--cf. Sec.~\ref{sec:system}).
After observing the environment, the agent should take an action.
In our text-focused model, this action is that of generating a new block of text--using TSL terminology, the action is an update to an output signal with a function term.
In particular, this function term is also a call to an LLM, asking for the next passage of the story (e.g. \texttt{toCave(summary)}--cf. Sec.~\ref{sec:system}).
Thus, for our work, a generative agent, $\mathcal{A} = (\sigma, \assign{\cdot})$, is a tuple of a strategy $\sigma$ and an implementation $\assign{\cdot}$ of the function and predicate terms ($\mathbb{F}$) provided in the specification.
The implementation $\assign{\cdot}$ in our case, corresponds to a particular version of an LLM.
We note that we also need to fix a random seed~\cite{LLMseed} in the LLM to ensure each generation is deterministic and is a pure function (the same input gives the same output every time) to match the model of TSL~\cite{finkbeiner2019temporal}.

A synthesized strategy realizing a TSL specification must be valid for \textit{any} implementation of the function and predicate terms.
This universal quantification of data manipulation is particularly well-suited for the context of LLMs, where the LLM behavior is treated as a black-box.
The correctness of synthesis is divorced from the inner-workings of the LLM, making this a well-suited approach to achieve procedural adherence and interpretability.

\section{System Overview}\label{sec:system}

Our proposed system combines reactive synthesis with an LLM, using reactive synthesis to control the high-level temporal reasoning while leaving the low-level details of each response to the LLM.
In this way, the reactive synthesis (and specifically the resulting automaton) controls the top level of the system, while LLM calls make up the computations that should happen at each transition step of the automaton.

We note that adhering to temporal properties (reactive synthesis) is in some ways fundamentally the same problem as probabilistic token-by-token generation (LLMs).
In both cases, we care about the relation of generated values in sequence.
The key contribution of the automaton as a mediating system for LLM generation is that, with TSL, we can directly encode our desired long-horizon temporal behavior.
In contrast, when using an LLM we must use prompt engineering to communicate our desired behavior in natural language--which is an error-prone, time-consuming task that still gives unreliable results.
Conceptually we expect this strategy to work well, as the structure of an automaton is relatively simple (even a few hundred states compares favorably to billions of parameters), lightening the load on the LLM.
By leveraging a hybrid of formal methods and LLMs, we can drive content generation in a way that allows the LLM to focus on a shorter context window.
% The specifications that generate the mediating automaton using reactive synthesis is a critical part of the system and usability design.
% In particular, with the temporal logic specification, we are able to deliver modularity to system developers.
% That is, an developer can add a new line of specification to their system and still be guaranteed that the system will preserve the behavior of the prior system as well as integrate the new behavior.

In Fig.~\ref{fig:overview}, we provide an overview of our system in action.
A single interaction in a conversation with the generative agent looks like a traditional turn from the end-user's perspective, but a four-step process is running under-the-hood each time they prompt the agent for a new passage: 
\begin{enumerate}
\item An evaluator LLM inspects the previous turn and outputs a series truth values (or context) describing events that did or did not occur during that turn. 
\item The automaton takes the responses from the evaluator LLM as input and, informed by its current state, selects a corresponding end-user prompt modifier so that the agent's response aligns with it's formal behavior. 
\item The end-user's prompt, the prompt-modifier, and a summary of previous passages are fed to a generator LLM. 
\item The generator LLM's response is outputted as the new passage to the end-user in the next time step and the process is repeated indefinitely. 
\end{enumerate}

In the context of the Fig.~\ref{fig:adventure} specification, the predicates \texttt{inCave()} and \texttt{inMarket()} are the queries that trigger the evaluator LLM. 
They query the evaluator for if, based on the previous story passage, the player is in a cave or a market. 
The function terms \texttt{toCave()} and \texttt{toMarket()} trigger the generator LLM.
They query the generator for the next passage using the end-user's prompt and a prescribed prompt modifier where the player visits a cave or a market. 

To ground the explanation of our system, we provide the concrete JavaScript code synthesized from the Fig~\ref{fig:adventure} specification in Appendix~\ref{app:code}, and example predicate and function term prompts in Appendix~\ref{app:prompts}.
% \markk{rewrite} We note that the automaton code does not exist in a vacuum. For the purposes of this work, we manually included implementations for predicates and functions like \texttt{toCave()} and \texttt{inCave()}, as well as the processing of automaton inputs and outputs to fully create the choose-your-own-adventure generator. We defer the task of generating predicate and function implementations, as well as domain-specific boilerplate code from the LLM to future work.

% \footnote{An implementation of our system in the context of a CYOA game can be found here: \url{https://barnard-pl-labs.github.io/CYOA-TSL/}}.

%\input{figs/overviewFig}

\section{Generative Agent Correctness}\label{sec:correctness}

Our architecture aims to address the aforementioned limitations of LLMs; adherence and interpretability. These properties are difficult to define formally for LLMs and their generated content~\cite{lipton2018mythos}.  
We again leverage the separation of data and control that TSL provides, and define these metrics ``up to'' the data level--taking a \textit{correctness modulo models} approach.
That is, we can provide a guarantee on the adherence and interpretability on the control level of the generative agents we create using automata synthesis, even if we we cannot guarantee these properties on the underlying models that are used as subcomponents in our specification.

\subsection{Adherence}\label{sec:adherence}

To provide a formal notion of an adherent generative agent, we introduce two complementary concepts of procedural adherence and point-in-time adherence. 
\subsubsection{Point-in-time adherence}
Point-in-time adherence captures the goal of self-consistency in the response to a prompt during one turn. 
That is, given a request for generative content, a self-reflection on the output of that request will match the intent.
For example, if a generative agent is prompted for a cave story passage, when the agent assesses its response, it will evaluate that the generated content is in a cave.
% put statistics crap here
%In the context of machine learning, this concept is most similar to computing the statistical probability...\raven{add LLM stuff here}.

Through the lens of TSL, point-in-time adherence is a property of the implementation $\assign{\cdot}$ of an agent $\mathcal{A}= (\sigma, \assign{\cdot})$.
Specifically, it is that, given a some abstract property $a$ (e.g. in a cave), and a function $f_a()$ that should generate content with that property, the output of that function can be checked by a complementary predicate $p_a()$. This is captured by the TSL formula: 
$\LTLglobally \left([cell \leftarrow f_{a}(...)] \LTLimplication \LTLnext p_{a}(cell)\right)$.
The point-in-time adherence of an agent $\mathcal{A}$ is measured by the point-in-time adherence of its function terms. 

% point in time adherence of an agent is measured by function term point in time adherence

\subsubsection{Procedural adherence}
In contrast to point-in-time adherence, procedural adherence captures the goal of self-consistency in the choice of the prompt itself during a multi-turn conversation.
That is, given a history of self-reflections (\textit{i.e.} predicate evaluations), a generative agent will attempt to generate a response (\textit{i.e.} select from a finite set of prompt modifiers) that matches it's formal specification of behavior. As an example, if an agent's specification dictates that during a game, the player should not visit a cave before a market, and the history of self-reflections over a particular game shows that the player has not yet visited a market, then the agent should not choose a cave prompt modifier to generate the next passage in the story.

Again, through the lens of TSL, procedural adherence is a property relating the strategy $\sigma$ of an agent $\mathcal{A}= (\sigma, \assign{\cdot})$ to the specification $\varphi$ from which it was synthesized.
Procedural adherence corresponds to the correctness of the synthesized automaton $\sigma$ from some specification $\varphi$.
% \begin{equation*} 
%   \branch{\sigma}{\iota}, \iota \sat \varphi
% \end{equation*}
The resulting agent $\mathcal{A}= (\sigma, \assign{\cdot})$, will have procedural adherence with respect to $\varphi$, regardless of the choice of $\assign{\cdot}$.

Procedural adherence lightens the load on the LLM by having the automaton serve as the \emph{memory} of the generative agent rather than the LLM itself; the automaton guides the conversation based on the context of the previous interactions with the end-user and leaves only the point-in-time generation decisions to the LLM. 
% two notions of time, interaction step and at each step theres the token generation steps
% given a history of interaction steps, we make the correct request, finite set of prompt modifiers, make the right chooice of prompt modifier at every time step, after you've vhosen the prompt, do you get the right response 
% we choose TSL as formal specification, regardless of form of specification, you have these two notions, seperate notions of correctness, seperately talking about these two things, different problems
% strategy is choosing from a finite set of prompt modifiers, conversational agents seperately from LLMs conversational agents, context is also LLM 
% sequence of tokens vs. sequence of prompts
% strategy is an infitine history of finite predicaet evaluations to choice of prompt in a finite space
% LLM token generation itself is the implementation 
% What is point in time adherence, statistics crap, for example, 

% Procedural adherence and point-in-time adherence are disjoint notions of adherence, and together would form full adherence. 

\subsection{Interpretability}\label{sec:interpretability}
To provide a formal notion of an interpretable generative agent, we introduce two complementary concepts of procedural interpretability and point-in-time interpretability in a similar fashion. Causality is commonly used to formalize one aspect of the interpretability of ML models~\cite{black2021consistent,verma2020counterfactual,baier2021verification}--a system is interpretable when causality of its output can be identified.
\subsubsection{Point-in-time interpretability}
Point-in-time interpretability seeks to reason about generative agent causality using individual tokens of a prompt and response after one turn. A token (or substring) in a prompt can be identified as a cause for the agent's response if, in changing that token, the agent generates a contextually different response. 

In machine learning, this concept is most similar to computing input element importance scores and providing saliency-based visualizations like heatmaps or highlighting to visualize relationships between inputs and outputs~\cite{li-etal-2019-cnm, bahdanau2014neural}. 
In the context of LLMs and NLP, related work seeks to address this lower level of interpretability by reasoning over knowledge graphs~\cite{luo2023reasoning},  and recent experimental work has achieved point-in-time interpretability through computing individual token or substring importance using Shapely values~\cite{goldshmidt2024tokenshap}.
By identifying high-importance tokens, developers can better understand which tokens are most likely to illicit a significantly different response from the LLM. 

While these methods are useful when determining the cause and effect of tokens over a single interaction between the LLM and an end-user, they fall short in reasoning about the causality of LLM output at the turn-level of abstraction and when working with a potentially infinite history of turns. 

\subsubsection{Procedural interpretability} 
Procedural interpretability seeks to reason about generative agent causality using a history of self-reflections and prompt choices after a multi-turn conversation. In the context of TSL, we refer to the work on temporal causality for reactive systems by Coenen et al.~\cite{coenen2022temporal}, which builds upon the formalization of causality from Halpern and Pearl~\cite{Pearl2005} by extending the definition of causality to reactive systems that run on infinite streams of input.
Core to the notion of causality is that ``a cause is an event such that, if it had not happened, the effect would not have happened either"~\cite{coenen2022temporal}.
We remark that the definition of an event itself is critical to the understanding of interpretability in our system.
In our system, the strategy~$\sigma$ of an agent $\mathcal{A}= (\sigma, \assign{\cdot})$ is procedurally interpretable because for any given choice of an update term, we can identify a predicate evaluation (or a set of) such that if the predicate evaluation had been different, we would have made a different choice of the update term. 

To demonstrate our notion of procedural interpretability, we turn to recent work from Finkbeiner et al. on synthesizing temporal causes~\cite{finkbeiner2024synthesis}. Finkbeiner et al. present a complete cause-synthesis algorithm, CORP\footnote{CORP tool: \url{https://github.com/reactive-systems/corp}} such that given a reactive system, a trace $\pi$, and an effect $\bm{\mathsf{E}}$, CORP can compute the closest cause $\bm{\mathsf{C}}$ for $\bm{\mathsf{E}}$. Colloquially, a trace is an infinite sequence of inputs and outputs over an execution of the system, a cause is a trace property reasoning only over the system's input variables, and an effect is a trace property of ranging only over its output variables. 
%For a full explanation of the cause-synthesis algorithm, we refer the reader to~\cite{finkbeiner2024synthesis}, as it is not critical to understanding this example.

In the context of choose-your-own-adventure games, take for example the generative agent described by the specification in Fig~\ref{fig:adventure} and implemented in Fig.~\ref{fig:code}. In summary, the agent is attempting to constrain the story so that the player must visit a market before visiting a cave. For the sake of this example, let us say that the developer working on this agent wants to reason about why, over a particular game, the player has not visited a cave yet. The definitions of this effect, the particular game trace, and two representations of the computed cause are depicted in Fig.~\ref{fig:cause}.
\begin{figure}[h]
\centering
% \vspace{-1.5em}
    \begin{align*}
    & \pi = \{\neg inMarket, \neg inCave, \upd{passage}{toMarket} \}^\omega \\[0em]
    & \varphi_E = \LTLglobally \neg \upd{passage}{toCave}\\[0em]
    & \varphi_C = \LTLglobally \neg inMarket
    \end{align*}
    $\mathcal{A}_C:$\begin{tikzpicture}
    \node[state, initial] (q1) {};
    \node[state, accepting, right of=q1] (q2) {};
    \draw   (q1) edge[above, ->, >=stealth] node{$\neg inCave \land \neg inMarket$} (q2)
            (q2) edge[loop right, >=stealth] node{$\neg inMarket$} (q2);
\end{tikzpicture}
    \caption{Effect $\varphi_E$, trace $\pi$, cause automaton $\mathcal{A}_C$, and manually guessed cause $\varphi_C$ given the reactive system synthesized from Fig.~\ref{fig:adventure}.}
    \label{fig:cause}
\end{figure}
% \vspace{-0.9em}

The developer has logged the sequence of inputs and outputs of the agent's internal automaton over this game, which can be described by the trace $\pi$. The effect $\bm{\mathsf{E}}$ can be described by the TSL formula $\varphi_E$ which states that the automaton never prompts the LLM for a passage where the player visit a cave. Procedural interpretability allows the developer to formally reason about the cause of automaton outputs--or update terms in the context of TSL--prompting the LLM for certain story passages, even if they cannot reason about the cause of LLM outputs.

CORP returns the cause $\bm{\mathsf{C}}$ in the form of a nondeterministic Büchi automaton $\mathcal{A}_C$~\cite{buchi1962decision}, visualized in Fig.~\ref{fig:cause}. By looking at this automaton, the developer can infer that the reason why the system has not prompted the LLM for a cave passage is because the player has not yet visited a market, represented by the TSL formula $\varphi_C$. We know that we can indeed always identify such a cause for our generative agent because Coenen et al. show that causality for reactive systems is decidable~\cite{coenen2022temporal}.

%One could imagine a fully integrated system where developers can design their generative agents, and use an embedded tool like CORP to gain an automated procedural interpretability of their agent.  

\section{Experiments}\label{sec:eval}
In our preliminary experiments, we evaluate the efficacy of imposing temporal constraints on generative agents with Temporal Stream Logic (TSL) to demonstrate the benefits of procedural adherence, focusing on agents that generate choose-your-own-adventure games (cf. Figs.~\ref{fig:overview} and~\ref{fig:adventure}).

\subsection{Methods}\label{sec:methods}
\label{sec:studyDesign}
We measured how well the choose-your-own-adventure agent adhered to three narrative constraints written in TSL and a natural language equivalent. The three tasks that the agent had to follow are shown below and the full NL prompts and TSL specs can be found in Appendix~\ref{app:specs}.
\begin{enumerate}
\item \emph{``The player can visit a cave only after visiting a town and a market."}
\item \emph{``The initial story setting must be a forest."}
\item \emph{``If the player makes three safe choices along their journey and they are in a cave, they must visit a town."} 
\end{enumerate}

Constraints 1 and 2 resemble the Fig~\ref{fig:adventure} spec and serve as baseline comparisons for the automaton-enhanced and pure-LLM agents. Constraint 3 evaluates how well the agents adhere to their guarantees for more complicated systems. 

We implemented both the automaton-enhanced agent outlined in Fig~\ref{fig:overview} and the pure-LLM agent in a JavaScript interface querying the GPT-4 API. At each time-step, the pure-LLM agent was prompted to generate the next passage of the story based on a story summary, a narrative constraint (\textit{e.g.} ``visit market before cave"), and the end-user input guiding the story (\textit{e.g.} ``take left path"). 
We played 75 games for 20 turns (or story passages) with each agent. 
For each agent, we then analyzed the percentage of games that resulted in hallucinations and/or arithmetic errors vs. games where all guarantees were satisfied for all turns. If a guarantee was violated in one turn in a game (\textit{e.g.} if the player visited a cave before a market or town) the whole game was marked as a violating game. 
We chose 75 games due to budget constraints.

\subsection{Results}

\begin{table}[t]
    \centering
    \renewcommand{\arraystretch}{1.2}
    \small 
     %\markk{Brian suggested stacked bar plots here - or just plot changes in bad hallucinations (3x reduction)}
\setlength{\tabcolsep}{7pt}

    \begin{tabular}{>{\centering\arraybackslash}p{0.5cm}p{2.5cm}>{\centering\arraybackslash}p{0.8cm}c}
        \toprule
        \textbf{Task} & \textbf{Method} & \textbf{States} & \textbf{Adherence (\%)} \\
        \midrule
        \multirow{2}{*}{1} & \textbf{adventure.tsl} & 8 & 96.00\% \\
        & pure LLM & - & 86.60\% \\
        \midrule
        \multirow{2}{*}{2} & \textbf{adventure.tsl $\land$ \newline 
 forest.tsl} & 9 & 96.00\% \\
        & pure LLM & - & 89.33\% \\ 
        \midrule
        \multirow{2}{*}{3} & \textbf{adventure.tsl $\land$ \newline choices.tsl} & 16 & 98.67\% \\
        & pure LLM & - & 14.67\% \\
        \midrule
        \multirow{2}{*}{4} & \textbf{adventure.tsl $\land$ \newline choices.tsl $\land$ \newline forest.tsl} & 17 & 97.33\% \\
        & pure LLM & - & 16.00\% \\ 
        \bottomrule
    \end{tabular}
    \caption{Results of four different tasks constraining generative agent behavior, evaluated over 75 games.}
    \label{tab:cyoa_results}
\end{table}

\begin{table}[t]
\setlength{\tabcolsep}{3pt}
    \centering
    \renewcommand{\arraystretch}{1.2}
    \small 
\begin{tabular}{lcccc}
        \toprule
        & \multicolumn{1}{c}{\textbf{Method}} & \multicolumn{2}{c}{\textbf{Violating Games}} & {\textbf{Total Games}} \\
        \cmidrule(lr{1em}){3-4}
        & & \textbf{Hallucinations} & \textbf{Arithmetic Errors} & \\
        \midrule
        & TSL & 9 (3.00\%) & 0 & 300 \\
        & LLM & 43 (14.33\%) & 101 (33.67\%) & 300 \\
        \bottomrule
    \end{tabular}
    \caption{Hallucinations and arithmetic errors over total specification violations across all games.}
    \label{tab:violation_results}
\end{table}

Table~\ref{tab:cyoa_results} shows an evaluation of our approach on the storytelling agent described in Sec.~\ref{sec:studyDesign}. For each task, the automaton-enhanced agent achieved high adherence, meaning that the agent generated passages that consistently behaved according to its specified long-horizon behavior. Our adherence measurements capture both procedural and point-in-time adherence for the automaton-enhanced agents, whereas we can only measure the point-in-time adherence of the pure-LLM agents. We note that the automaton-enhanced agents all achieve 100\% procedural adherence by construction, and the lack of point-in-time adherence leads to LLM hallucinations and specification violations. These results show that the introduction of procedural adherence to a generative agent lightens the load on the LLM by handling the context and memory of the interaction. 
% Procedural adherence leads to a decrease in LLM hallucinations overall, and standardizes the frequency of LLM hallucinations   

%For Task 1 and 2, the pure-LLM agent also did fairly well at adhering to its natural language specifications and only hallucinated a cave or failed to generate a forest on the first passage 13.4\% and 10.67\% of the time. Task 3 involves the combination of two distinct specifications, which constrains the locations the player visits during the game and the impact of their choices on gameplay. Task 4 is the combination of all three specifications outlined in Sec~\ref{sec:methods}. Combining two TSL specifications that impact different aspects of the gameplay had no observable effect on how well the automaton-enhanced agent satisfied its guarantees. In contrast, the pure LLM-based agent encountered challenges when tasked with managing multiple objectives. In Task 3, the agent demonstrated an increased tendency to hallucinate a cave in approximately 21\% of the games, compared to 13\% in Task 1. This 8\% rise in cave hallucinations illustrates that without any high-level structure to natural language prompts, as they become more verbose, the consistency of the generated responses diminish. 

Table~\ref{tab:violation_results} highlights another finding: the pure LLM-based agent made simple arithmetic errors more than twice as often as hallucinations. 
We classify arithmetic errors as violations of the \texttt{choices.tsl} spec, when the LLM inaccurately updated the count of safe choices the player had made during their game, either by inconsistently increasing the count or doing so when the player was not in a town. 
We classify hallucinations as violations of either the \texttt{forest.tsl} or \texttt{adventure.tsl} specs, when the LLM generated a passage in a cave before visiting both the town and a market, or when the first passage was not set in a forest. 
It is worth noting that LLMs often encounter challenges when generating responses to numerical and mathematical queries, and this weakness has been deemed an research area of critical importance~\cite{yen2023three}. TSL and reactive synthesis offer a formally correct approach.

One limitation of this evaluation is the quality of the natural language prompt we gave the pure LLM-based agent and its equivalency to the TSL specification of the automaton-enhanced agent. However, it is worth noting that it took us less time to engineer all TSL specifications than it took to engineer satisfactory natural language prompts to ensure semi-consistent behavior from the pure LLM-based agent. 

Overall, our preliminary evaluation suggests that guarantee of procedural adherence can effectively constrain long-horizon generative agent behavior in complex scenarios, outperforming pure LLM-based agent specifications especially when dealing with arithmetic computations.

\section{Discussion}
\subsubsection{Modularity}
The automaton-enhanced agent's performance in Tasks 3 and 4 also demonstrate the benefit of using program synthesis to gain procedural adherence. Using TSL to synthesize the generative agent's underlying automata adds a level of \emph{modularity} to the system. A modular generative agent in this case means that given two or more realizable TSL specifications describing different generative agents, we can compose--or logically $\land$--them to get a new specification that, in theory, captures the guarantees of the original component specifications. 
When running synthesis on this new specification, we could get one of two results: 
\begin{enumerate}
\item \textbf{Realizable}; we can synthesize a new control strategy for a generative agent with the same procedural adherence guarantees as agents synthesized from the component specs, as is the case for specs in Tasks 2-4.
\item \textbf{Unrealizable}; there is no control strategy that satisfies all of the constraints of the new spec.
\end{enumerate}

An unrealizable spec results from a conflict between the constraints of the component specs. 
If two developers are collaborating on creating a generative agent and have conflicting requirements for agent behavior, an LLM would take these requirements and produce some output without any guarantee of alerting the developers to the conflict. 
This modularity is notoriously difficult to achieve with natural language prompt engineering~\cite{cheng2023prompt}.
In contrast to a pure-LLM system, synthesis frameworks are able to produce explanations for what can go wrong and suggestions for how to modify the specification to make it achievable. 
Furthermore, they can provide feedback about inconsistencies and redundancies. 
There are tools to help developers debug their unrealizable specifications, such as counter-strategies and error localization~\cite{cui2024towards, li2011mining,konighofer2013debugging,alur2015pattern}. 

To enable an iterative creation process, system developers can build upon existing versions of their specifications by modifying or adding one component at a time. This allows developers to break their ultimate goal into smaller sub-goals in making sure that each iteration produces a functional version of their desired program. The modular design of TSL also supports re-using of specifications, allowing the developer to scale successful design patterns.
We note that while we could get procedural adherence and interpretability from manual automaton construction (as opposed to using TSL synthesis), it is TSL as a specification language and the synthesis process that gives us modularity. 

\section{Threats to Validity}
Some threats to the internal validity of our experiments include the impact of a limited budget and further prompt engineering. 
Each task could have been evaluated over more games to allow for broader assertions about LLM behavior and hallucinations. 
Additionally, we could have improved the prompt engineering and paired the generative agent with a monitor to check and regenerate violating responses before presenting them to the end-user, as done by~\cite{yang2023plug}. 
To partially mitigate these threats, we chose to use GPT-4 to compare against the state-of-the-art at the time. 
%We also recognize that prior work has raised concerns over pre-trained LLM performance evaluations as a whole due to the lack of transparency and complexity of the models~\cite{ouyang2023llm}.

One challenge to the wider adoption of our work, and thus a key threat to external validity, is the documented difficulty of temporal logic usability~\cite{jayagopal2022exploring,rothkopf2023towards,de2007user,yamashita2017visual}. 
There is still a barrier to writing TSL and other specification languages for program synthesis, as they do not share syntax with common programming languages and are semantically complex. One way to solve this is to provide better tools for writing specifications~\cite{rothkopf2023towards}. 
We reiterate that TSL is one of many possible ways of discussing and achieving procedural adherence and interpretability in generative agent behavior.
%, and while its important to combat these usability hurdles, they do not invalidate our framework as a whole. 

% Furthermore, the efficacy of our framework is contingent upon its design, rendering it more suited for certain tasks than others. Since TSL introduces temporal logic, users must articulate their intentions within the framework of TSL semantics. Consequently, tasks involving temporal operations or possessing sequential properties stand to gain greater advantages from the incorporation of temporal logic.

% A threat to the construct validity of our work is the indeterminacy of point-in-time adherence and interpretability. For our purposes, we formalize \emph{procedural} adherence and interpretability in Secs~\ref{sec:adherence} and~\ref{sec:interpretability} as they pertain to our definition of generative agents, but their complementary \textit{point-in-time} concepts remain undefined in the NLP field.
% Despite these limitations, our results serve as a targeted case study and proof of concept for the combination of formal methods and LLM generation to bring procedural adherence and interpretability to generative agents.

\section{Conclusion}

This paper combines formal logic-based program synthesis with large language model (LLM) content generation to address key challenges in long-horizon interactions with pre-trained LLMs. Leveraging formal models, system developers can enforce constraints on LLM-generated content to create more adherent and interpretable autonomous agents. 

% In an evaluation of our framework, we demonstrated the efficacy of procedural adherence in constraining the long-horizon behavior of a choose-your-own-adventure game agent through a set of predefined tasks. 

% While achieving true point-in-time adherence and interpretability in the age of LLMs remains a critical and open question in NLP and the explainable AI community, we provide procedural adherence and interpretability using formal methods to narrow the scope of the problem.

% For future work, we plan to evaluate the benefits of the procedural properties of autonomous interactive agents using human participants. There is an increasing need for safer, more transparent agents that remain flexible to real-world inputs; we hope that our results provide more motivation for the combination of formal models and LLM content generation to meet this need. 

% Our architecture offers a higher level control over coherency in generative agent behavior up to the prompt. The introduction of procedural interpretabiliity in the system allows engineers to comprehend and explain agent decision-making processes. The iterative creation process facilitated by our architecture empowers system engineers to identify conflict, refine, and experiment with their specifications incrementally.

% \subsection*{Acknowledgements}
% This material is based upon work supported by the National
% Science Foundation under Grant No. CCF-2105208.

\bibliography{refs}

\newpage
\appendix
\section*{Technical Appendix}
\section{Code}\label{app:code}
We provide the concrete JavaScript code synthesized from the motivating Fig~\ref{fig:adventure} specification that serves as central automaton of our generative agent implementation.
\begin{figure}[h]

\begin{lstlisting}[style=JavaScript]
if (currentState === 0) {
  if (!inCave(s) && !inMarket(s)) {
    storyPassage = toMarket(s)
    currentState = 4 } }
else if (currentState === 1) {
  if (!inCave(s)) {
    storyPassage = toCave(s)
    currentState = 3 } }
...
else if (currentState === 3) {
  if (inCave(s) && inMarket(s)) {
    storyPassage = toMarket(s)
    currentState = 1 }
  ...
  else if (inCave(s) && !inMarket(s)) {
    storyPassage = toMarket(s)
    currentState = 4 } }
else if (currentState === 4) {
  if (!inCave(s) && inMarket(s)) {
    storyPassage = toMarket(s)
    currentState = 1 }
  ...
  else if (!inCave(s) && !inMarket(s)) {
    storyPassage = toMarket(s)
    currentState = 4 } }
\end{lstlisting}

% \begin{lstlisting}[style=JavaScript]
% if (currentState === 0) {
%   if (!inCave(summary) && !inMarket(summary) && !inTown(summary)) {
%     storyPassage = toTown(summary)
%     currentState = 1
%   }
%   ...
% }
% else if (currentState === 1) {
%   ...
%   else if (!inCave(summary) && inMarket(summary) && !inTown(summary)) {
%     storyPassage = toTown(summary)
%     currentState = 3
%   }
%   ...
% }
% ...
% else if (currentState === 3) {
%   ...
%   else if (!inCave(summary) && inMarket(summary) && inTown(summary)) {
%     storyPassage = toTown(summary)
%     currentState = 4
%   }
%   ...
% }
% else if (currentState === 4) {
%   ...
%   else if (!inCave(summary)) {
%     storyPassage = toCave(summary)
%     currentState = 5
%   }
% }
% ...
% \end{lstlisting}
\caption{A snippet of the JavaScript code synthesized from the   Fig~\ref{fig:adventure} TSL spec representing a 5-state Mealy machine. The story \texttt{summary} has been abbreviated to \texttt{s}.}
\label{fig:code}
\end{figure}
The automaton is a straightforward sequence of nested conditionals with a global \texttt{currentState} variable. 
As input, the automaton receives predicate evaluations (e.g. \texttt{inCave(summary)}) from the evaluator LLM with information about the locations the player has visited in the story so far.
As output, the automaton queries the generator LLM for the next appropriate location (e.g. \texttt{toCave(summary)}) and updates the \texttt{storyPassage} with LLM's response as it transitions to the next state. 
After each transition, the \texttt{summary} is updated with the newly generated \texttt{storyPassage}, as well as the \texttt{currentState}.

\section{Predicate and Function Term Prompts}\label{app:prompts}
We provide example GPT-4 prompts representing function and predicate terms seen in Appendix \ref{app:code}. 

Function term prompt for \texttt{toCave(summary)}:

\vspace{5pt}
\noindent\emph{``You are writing a choose your own adventure book. Compose a one paragraph-long passage of the story of at most 100 words. The paragraph should end just before a critical choice. Do not specify choices. Write in the present tense. Compose a passage where the reader encounters a \textbf{cave} on their journey."}

\vspace{5pt}
Predicate evaluation prompt for \texttt{inCave(summary)}: 

\vspace{5pt}
\noindent \emph{``Read this passage in an adventure story. Is the main character actively in a \textbf{cave} or not? Respond '0' if it is false, or '1' if it is true."}

\vspace{5pt}
New story \texttt{summary} generation prompt: 

\vspace{5pt}
\noindent\emph{``You are writing a book and need to recall important points of the story so far. Summarize the provided passage from about the story so far in moderate detail, including the main character description, the locations visited, items acquired, and interactions with other characters."}

\section{Specifications}
\label{app:specs}
We provide the full list of the concrete TSL specifications and natural language prompts corresponding to each task below. For readability, we shorten the name of the \texttt{summary} variable seen in earlier specs and code to \texttt{s}. 

\subsection{Task 1}
\subsubsection{adventure.tsl}
spec
\begin{lstlisting}[basicstyle=\ttfamily\footnotesize]
initially assume {
    ! inCave(s);
    ! inMarket(s);
    ! inTown(s);    
}

always assume {
    [storyPassage <- toCave(s)] <-> X inCave(s);
    [storyPassage <- toMarket(s)] -> F inMarket(s);
    [storyPassage <- toTown(s)] -> F inTown(s);
}

guarantee {
    (! (inCave(s))) W (inMarket(s));
    (! (inCave(s))) W (inTown(s));
}

always guarantee {
     F inCave(s);
     F inMarket(s);
     F inTown(s);
}
\end{lstlisting}
\subsubsection{Natural language prompt example: } ``You are writing a choose your own adventure book. Compose a one paragraph-long passage of the story of at most 100 words. The paragraph should end just before a critical choice. Do not specify choices. Write in the present tense. The player cannot visit a cave until they visit a town. The player cannot visit a cave until they visit a market. The player will eventually visit a town. The player will eventually visit a market. The player will eventually visit a cave. After the player visits a town and a market, they must visit a cave. The player may visit these locations in a later passage."
\subsection{Task 2}
\subsubsection{forest.tsl}
spec
\begin{lstlisting}
initially guarantee {
    [storyPassage <- toForest(s)]
}
\end{lstlisting}
\subsubsection{Natural language prompt example: }
``If this is the first passage of the story, you should write an introductory passage of the story starting which describes the character and the setting. The initial setting must be a forest, not a market, town or cave."

\subsection{Task 3}
\subsubsection{adventure.tsl \&\& choices.tsl}
specs
\begin{lstlisting}[basicstyle=\ttfamily\footnotesize]
initially assume { 
    ! safeThreshold;
    ! inCave(s);
    ! inTown(s);
    ! inMarket(s);
}

always assume {
    [safeCount <- safeCount + 1] -> F safeThreshold;
    [storyPassage <- toCave(s)] <-> X inCave(s);
    [storyPassage <- toTown(s)] -> F inTown(s);
    [storyPassage <- toMarket(s)] -> F inMarket(s);
}

guarantee {
    (safeThreshold && inCave(s) -> F inTown(s);
    (! (inCave(s))) W (inTown(s));
    (! (inCave(s))) W (inMarket(s));
}

always guarantee {
    (safe && inTown(s)) <-> 
        [safeCount <- safeCount + 1];
     F inCave(s);
     F inTown(s);
     F inMarket(s);
}
\end{lstlisting}
\subsubsection{Natural language prompt example: }
``You are writing a choose your own adventure book. Compose a one paragraph-long passage of the story of at most 100 words. The paragraph should end just before a critical choice. Do not specify choices. Write in the present tense. The player cannot visit a cave until they visit a town. The player cannot visit a cave until they visit a market. The player will eventually visit a town. The player will eventually visit a market. The player will eventually visit a cave. After the player visits a town and a market, they must visit a cave. The player may visit these locations in a later passage. If the player is in the town and makes a safe choice, the `safeCount` count variable should be incremented by 1. The current amount of safe choices made is: 0. The the player just made a neutral choice. Once the player has made 3 safe choices and they are in a cave, they should go to a town. Make sure to output the current amount of safe choices made in the form: `safeCount=x` at the beginning of each story passage and then include the story, but remember that this variable can only be updated if the player is in a town, which right now, is false."   

\end{document}